\documentclass{article}
\usepackage[preprint]{neurips_2026}

\usepackage[utf8]{inputenc}
\usepackage[T1]{fontenc}
\usepackage{hyperref}
\usepackage{url}
\usepackage{booktabs}
\usepackage{amsfonts}
\usepackage{amsmath}
\usepackage{amssymb}
\usepackage{graphicx}
\usepackage{multirow}
\usepackage{xcolor}

\title{Do Better Imagined Rollouts Mean Better Robot Control? A Controlled Study of World-Model Evaluation Under Feedback}

\author{
  Dharini Raghavan, Amritpal Singh \\
  Georgia Institute of Technology, Emory University 
}

\begin{document}

\maketitle

\begin{abstract}

Predictive models are increasingly used in robotics for state estimation, planning, control, and policy evaluation, yet they are often judged by open-loop prediction accuracy over a fixed horizon. In closed-loop operation, a robot repeatedly acts, receives new measurements, updates its state estimate, and recomputes control. We study this difference in a differential-drive path-tracking task with biased odometry and intermittent landmark sensing. Six state estimators are evaluated across 24 sensing conditions using trajectory replay, a 20-step measurement-free rollout, and closed-loop tracking. Replay position RMSE correlates more strongly with closed-loop cross-track RMSE than rollout error (Spearman rho = 0.923 vs. 0.774) and selects a different estimator from the closed-loop optimum in 5/24 conditions, compared with 18/24 for the rollout metric. We then vary rollout horizon and measurement-update interval. With H=20, rank agreement decreases from rho = 0.916 with measurements at every step to rho = 0.774 with no measurements. A horizon-update grid shows that long prediction horizons remain informative when regular corrections are retained, whereas long rollouts without correction can produce rankings that differ substantially from closed-loop behavior. We also test recurrent estimators trained on longer sensing outages. This improves the EKF-anchored models under combined sensing degradation, reducing GRU-EKF cross-track RMSE from 1.72 m to 1.06 m, but the gain is not consistent across isolated outages or estimator architectures. These results show that predictive-model evaluation in robotics should specify both prediction horizon and measurement-update schedule. For models used in feedback, offline rollouts are most informative when their sensing and correction pattern reflects closed-loop operation. Code is available at \url{
https://github.com/rdharini2001/Robot_World_Model}

\end{abstract}


\section{Introduction}

Predictive models are becoming an increasingly important component of modern robotic systems. They are used to estimate future states, support planning, evaluate candidate control actions, and, in some recent systems, generate control commands directly from learned dynamics or video representations \citep{agarwal2025cosmos,gemini2025veo,ye2026dreamzero,kim2026cosmospolicy}. As these models take on a larger role in the perception-planning-control loop, their evaluation becomes increasingly important. In particular, it is necessary to understand which measures of prediction quality are actually informative about the behavior of the complete robotic system.

Most predictive models are evaluated using an offline measure of fidelity, such as state-prediction error, reconstruction error, perceptual similarity, or error accumulated over a multi-step rollout. These quantities are useful because they measure how closely the predicted trajectory follows the underlying system. In robotics, however, prediction rarely occurs in isolation. A controller computes an action from the current state estimate, the robot moves, new sensor measurements become available, the state estimate is updated, and a new control command is computed. Prediction error therefore affects both the current control action and the future trajectory from which subsequent measurements are obtained. The resulting closed-loop behavior depends not only on the magnitude of the prediction error, but also on when that error occurs, how it affects the controller, and when new measurements become available to correct it.

This distinction is particularly important when predictive models are evaluated through multi-step rollouts. A long observation-free rollout provides a useful measure of how well a model can propagate its state without external correction, and such an evaluation resembles the prediction stage used in model-based planning. Many robotic systems, however, receive measurements and recompute control inputs at a much higher frequency than the length of a typical evaluation rollout. A model that accumulates error during an extended open-loop prediction may still perform well when regular sensor updates are available. Conversely, a model with low rollout error may produce state errors that are especially detrimental to the feedback controller. The relationship between rollout accuracy and closed-loop performance therefore depends on the sensing and update schedule under which the model is used.

Studying this relationship directly in large visual world models is difficult because several effects occur simultaneously. Representation error, learned dynamics, action conditioning, observation frequency, policy behavior, and replanning all influence the resulting trajectory. We therefore study the problem in a controlled mobile-robot setting in which these factors can be separated. A differential-drive robot follows a reference path using a feedback controller while receiving biased wheel odometry and intermittent range-bearing measurements from known landmarks. The controller operates on the estimated pose rather than the ground-truth pose. Estimation error consequently changes the applied control input and, through the robot dynamics, changes the subsequent trajectory and sensing geometry.

Within this setting, we use state estimators as compact predictive models of robot motion. Each estimator propagates an internal state using odometry and incorporates landmark observations when they are available. The estimator can also be propagated forward for several steps after landmark measurements are removed, allowing the same model to be evaluated under both measurement-corrected and observation-free prediction. This formulation removes the additional complexity of image generation and learned visual representations while retaining the feedback structure that is central to robotic deployment. It therefore provides a controlled setting for asking how offline prediction metrics relate to closed-loop control when the state estimate is periodically corrected by new measurements.

We evaluate six estimators: dead reckoning (DR), an extended Kalman filter (EKF), and four recurrent residual estimators that use either DR or the EKF as an analytic reference. The models are tested across 24 sensing conditions that vary landmark availability, range noise, gyroscope bias, and complete sensing outages. Each estimator is evaluated using three complementary protocols: replay on a common recorded trajectory, a 20-step measurement-free rollout initialized from the estimator's current internal state, and closed-loop path tracking in which the estimate directly drives the controller. Although the replay and rollout evaluations both measure state-prediction accuracy, they produce substantially different rankings of the estimators. Replay position RMSE has a Spearman correlation of $\rho=0.923$ with closed-loop cross-track RMSE and selects a different estimator from the closed-loop optimum in 5 of 24 conditions. The 20-step measurement-free rollout has a lower correlation of $\rho=0.774$ and selects a different estimator in 18 of 24 conditions.

We next separate the effects of prediction horizon and measurement-update frequency. We vary the interval between measurement corrections while holding the rollout horizon fixed, and we also evaluate a grid of rollout horizons and update intervals. The results show that ranking agreement with closed-loop performance decreases as measurement corrections become less frequent. At a fixed horizon of $H=20$, Spearman correlation decreases from $\rho=0.916$ when measurements are incorporated at every step to $\rho=0.774$ when no measurements are provided during the rollout. The horizon sweep shows that long rollouts can remain informative when regular measurement updates are retained, whereas long observation-free rollouts can produce rankings that differ substantially from those observed in feedback control. These experiments indicate that rollout horizon and measurement-update frequency should be considered together rather than treated as independent properties of an evaluation protocol.

We also examine the corresponding training question by exposing recurrent estimators to longer sensing outages during training. Longer-blackout training improves both EKF-anchored estimators under a combined sensing-degradation condition, including a reduction in GRU-EKF cross-track RMSE from $1.72$~m to $1.06$~m. The same training procedure does not provide uniform gains across isolated blackout conditions and does not improve the dead-reckoning-anchored model. A matched-data experiment, in which training-set size is held fixed while only the blackout distribution is changed, further shows that matching the training blackout distribution to the evaluation conditions does not by itself guarantee improved closed-loop behavior.

The experiments support three conclusions that are relevant to the evaluation of predictive models in robotics:
\begin{enumerate}
    \item \textbf{Prediction metrics should be evaluated under the sensing schedule of the intended closed-loop system.} A measurement-free rollout can provide a poor ranking of models when the deployed robot receives regular measurement corrections.

    \item \textbf{Rollout horizon alone does not determine whether an offline metric is informative for control.} The interval between measurement updates strongly affects the relationship between prediction error and closed-loop performance, particularly for longer prediction horizons.

    \item \textbf{Training for longer observation gaps does not produce uniform improvements in closed-loop robustness.} Its effect depends on the estimator structure and the sensing regime, even when the amount of training data is controlled.
\end{enumerate}

These results are not intended to suggest that short-horizon prediction is generally preferable to long-horizon prediction. Instead, they show that the usefulness of a rollout metric depends on the operating conditions under which the model will be used. For predictive models that operate inside a robotic feedback loop, the evaluation protocol should account for the timing of sensing, state correction, and control updates in addition to the prediction horizon itself.

\section{Related Work}

\textbf{World models for robot learning and control.}
World models provide a predictive representation of system dynamics that can be propagated forward to support planning, policy learning, and action evaluation \citep{ha2018worldmodels}. Dreamer and its subsequent variants use predicted latent-state trajectories to learn control policies from simulated experience generated by the model \citep{hafner2019dreamer,hafner2023dreamerv3}. More recent work has extended this idea to higher-dimensional visual observations and robotic manipulation. Cosmos develops video-based predictive models for physical systems \citep{agarwal2025cosmos}, DreamZero uses a learned world-action model to generate robot behavior without task-specific policy training \citep{ye2026dreamzero}, and Cosmos Policy adapts a video prediction model for visuomotor control and planning \citep{kim2026cosmospolicy}. Predictive models are also beginning to serve as evaluation environments. The Gemini Robotics evaluation framework, for example, uses an action-conditioned video simulator to compare robot policies under nominal and distribution-shifted conditions \citep{gemini2025veo}. Although these systems differ substantially in scale and representation, they share a common requirement: predictions must remain useful when they are coupled to repeated sensing, state updates, and control decisions. Our work studies this requirement in a lower-dimensional setting where the effects of prediction, measurement correction, and feedback can be examined separately.

\textbf{Prediction accuracy and closed-loop performance.}
The distinction between model accuracy and control performance has been studied extensively in model-based control and reinforcement learning. A model that accurately reproduces system trajectories does not necessarily provide the errors or sensitivities that matter most to a feedback controller. Lambert et al.~\citep{lambert2020objective}, for example, show that improved predictive accuracy of a learned dynamics model does not always lead to improved policy performance in model-based reinforcement learning. Our study examines a related issue at the state-estimation stage of the control loop. Rather than varying the dynamics model and retraining the controller, we keep the controller fixed and vary the estimator that supplies its state feedback. This allows us to measure directly how rankings based on offline state-prediction error correspond to rankings based on closed-loop path-tracking performance.

\textbf{Model-based and learned state estimation.}
Probabilistic state estimation remains a central component of mobile-robot localization and navigation. The extended Kalman filter (EKF) is widely used when the robot motion and sensor models are known but nonlinear \citep{thrun2005probabilistic}. More recent approaches combine this model-based structure with learned components in order to capture modeling errors, difficult observation processes, or residual dynamics that are not represented explicitly. Examples include Backprop KF \citep{ha2016backprop}, deep variational Bayes filters \citep{karl2017dvbf}, recurrent Kalman networks \citep{becker2019rkn}, differentiable particle filters \citep{jonschkowski2018dpf}, latent ordinary differential equation models \citep{rubanova2019latentode}, and KalmanNet \citep{revach2022kalmannet}. The recurrent estimators used in this paper follow the same general principle: an analytic state estimate is retained as a reference, while a learned recurrent model predicts a correction to that estimate. By comparing identical recurrent structures anchored to either dead reckoning or the EKF, we can also examine how strongly closed-loop robustness depends on the quality of the underlying model-based estimator.

\textbf{Robot state estimation from external sensing infrastructure.}
The state estimate used for feedback does not have to be obtained exclusively from sensors mounted on the robot. External sensing infrastructure can provide global observations of robot pose and can complement or replace portions of the onboard localization system. Raghavan et al.~\citep{raghavan2024zeroshot} present a zero-shot perception framework for estimating and tracking the pose of autonomous mobile robots using infrastructure-mounted vision sensors. Their results demonstrate how external cameras can provide a global estimate of robot motion without requiring the complete perception pipeline to reside onboard the platform. Zero-Splat TeleAssist~\citep{dokania2025zerosplat} extends this infrastructure-based viewpoint to semantic teleoperation, where observations from multiple commodity CCTV cameras are combined into a shared 6-DoF spatial representation for operating multiple robots. These studies illustrate that the state supplied to a planner or controller may be constructed from onboard sensing, external infrastructure, or a combination of the two. The question addressed in the present work is therefore complementary to the sensing architecture itself: given a state representation that will be used for feedback, how should its predictive accuracy be evaluated so that the resulting metric reflects closed-loop robot behavior?

\section{Controlled Experimental Testbed}
\label{sec:system}

We study the evaluation problem in a controlled planar mobile-robot setting. The robot state is
$s_t=(x_t,y_t,\theta_t)$, where $(x_t,y_t)$ denotes planar position and $\theta_t$ denotes heading. The vehicle follows standard unicycle kinematics with a sampling interval of $\Delta t=0.1$~s. A pure-pursuit path-tracking controller with look-ahead distance $L_d=0.8$~m is used to follow a lemniscate reference trajectory.

The robot receives two sources of state information. Wheel odometry provides incremental motion measurements, while range-bearing measurements to known landmarks provide intermittent absolute corrections. The odometry stream is corrupted by multiplicative wheel slip and additive gyroscope bias. Landmark sensing is degraded systematically along four axes: range-measurement noise, the number of visible landmarks, gyroscope bias, and scheduled sensing outages during which all landmark observations are unavailable. These perturbations allow us to study estimator behavior under gradually worsening sensing conditions while keeping the robot dynamics and control law fixed.

The controller does not have access to the ground-truth pose. Instead, each control input is computed from the pose estimate produced by the active observer. Estimation errors therefore affect the applied control command, which changes the subsequent robot trajectory and the measurements encountered later in the run. This coupling is central to the experimental design. An estimator that appears accurate when evaluated on a fixed trajectory may produce different behavior once its estimates are used directly in the feedback loop.

\subsection{Predictive State Models}

\textbf{Dead reckoning and extended Kalman filtering.}
Dead reckoning (DR) propagates the robot pose using odometry alone and does not apply measurement-based corrections. The extended Kalman filter (EKF) uses the same unicycle motion model together with analytic Jacobians for the landmark range-bearing observation model. Between landmark observations, the EKF propagates the state and covariance using the motion model; when a valid landmark measurement becomes available, it performs a standard measurement update.

\textbf{Recurrent residual observers.}
The learned observers retain either DR or the EKF as an analytic reference estimate and use a recurrent model to predict an additive residual correction. We consider two recurrent architectures: a gated recurrent unit (GRU) and a selective diagonal state-space model (SSM) \citep{gu2022s4,gu2024mamba}. Let $\bar{s}_t$ denote the pose produced by the analytic estimator and let $r_t$ denote the learned residual. The corrected pose estimate is

\[
    \hat{s}_t = \bar{s}_t \oplus r_t,
\]

where $\oplus$ denotes pose composition with the predicted residual correction. The six primary observers considered in the study are DR, EKF, GRU-DR, SSM-DR, GRU-EKF, and SSM-EKF. The recurrent models are trained by backpropagation through time using 200 domain-randomized expert trajectories that span variations in sensing quality and motion uncertainty.

\textbf{Training with extended sensing outages.}
To examine whether exposure to longer periods without landmark measurements improves robustness, we train additional versions of SSM-EKF, GRU-EKF, and SSM-DR, denoted by $\dagger$. These models are trained on 300 trajectories, half of which contain an additional sensing outage lasting between 32 and 58 time steps. The standard training data contain outages lasting between 6 and 27 steps. The network architecture, training objective, and optimizer are otherwise unchanged.

The sensing outage is introduced by masking the landmark-measurement channels while leaving the simulated vehicle dynamics and odometry sequence unchanged. Because landmark sensing does not alter the physical evolution of the simulated robot, this procedure isolates the effect of missing measurement updates while preserving the underlying motion. The resulting models allow us to test whether training on longer periods of prediction without external correction translates to improved closed-loop performance.

\section{Evaluation Protocols}
\label{sec:evaluation}

We evaluate each observer under three protocols that differ primarily in the availability of future measurements and in whether the estimated state is allowed to influence the robot trajectory. Using the same set of observers under all three protocols allows us to compare conventional offline estimation accuracy, measurement-free prediction accuracy, and actual closed-loop control performance.

\textbf{Trajectory replay with measurement updates.}
In the replay evaluation, every observer processes the same recorded trajectory. The input sequence contains the realized odometry, landmark range-bearing measurements, and the corresponding measurement-availability masks. Because the robot trajectory is fixed, differences between observers do not alter the future inputs presented to the estimator. We evaluate the resulting pose estimates against the ground-truth trajectory using position RMSE and heading RMSE. This protocol represents the conventional offline evaluation of localization or state-estimation accuracy.

\textbf{Measurement-free rollout.}
The second protocol evaluates the ability of each observer to propagate its state estimate in the absence of future landmark corrections. We first replay the recorded trajectory through the observer and save its complete internal state at regularly spaced checkpoints. At each checkpoint, a copy of the observer is propagated forward for $H=20$ steps using only the realized odometry sequence. Landmark measurements are withheld for the entire rollout. Position RMSE is measured at the final rollout horizon and aggregated across checkpoints and ten evaluation seeds.

This evaluation measures how rapidly each observer accumulates state error when it must rely entirely on motion propagation for two seconds. Unlike the replay protocol, the measurement-free rollout explicitly tests multi-step prediction without external correction, while still using the same realized odometry sequence for every observer.

\textbf{Closed-loop path tracking.}
In the closed-loop evaluation, the observer operates online inside the simulator and supplies the pose estimate used by the pure-pursuit controller. The resulting control commands are applied to the simulated robot, so estimator errors directly affect the subsequent vehicle trajectory. Landmark measurements continue to arrive according to the sensing condition being evaluated, and the observer incorporates those measurements whenever they are available. We use cross-track RMSE as the primary measure of path-tracking performance and additionally record control effort, recovery time following sensing outages, and the fraction of runs that diverge.

The three protocols therefore expose each observer to different information and feedback structures. Replay permits repeated measurement correction but holds the robot trajectory fixed. The measurement-free rollout removes landmark corrections for a prescribed prediction horizon while still replaying the realized odometry sequence. Closed-loop evaluation combines intermittent measurement updates with feedback, allowing estimation errors to alter future control inputs, robot motion, and sensing geometry. The central question of the paper is whether either offline protocol preserves the ranking of observers obtained when those same observers are used for closed-loop control.

For each of the 24 sensing conditions, every offline metric produces a ranking over the six primary observers. We compare this ranking with the ordering obtained from closed-loop cross-track RMSE under the same sensing condition. In addition to the overall Spearman rank correlation across the 144 observer-condition pairs, we report how often an offline criterion selects an observer other than the closed-loop optimum and quantify the resulting selection regret relative to the best closed-loop observer for that condition.

\section{Results}

\subsection{The analytic estimator provides the main source of nominal closed-loop stability}
\label{sec:baseline-results}

\begin{table}[t]
\centering
\small
\caption{Nominal performance averaged over ten evaluation seeds. Learned residual correction provides the largest benefit when it is applied on top of the EKF estimate.}
\label{tab:nominal}
\begin{tabular}{lccc}
\toprule
Observer & Replay pos. RMSE (m) & Cross-track RMSE (m) & Divergence fraction \\
\midrule
DR             & 1.503 & 1.045 & 0.260 \\
EKF            & 0.140 & 0.212 & 0.004 \\
GRU-DR         & 0.281 & 0.694 & 0.140 \\
SSM-DR         & 0.325 & 0.909 & 0.196 \\
GRU-EKF        & \textbf{0.094} & \textbf{0.163} & 0.000 \\
SSM-EKF        & 0.121 & 0.199 & 0.000 \\
\bottomrule
\end{tabular}
\end{table}

Table~\ref{tab:nominal} summarizes estimator performance under nominal sensing conditions. GRU-EKF achieves both the lowest replay position error and the lowest closed-loop cross-track error. SSM-EKF also performs well, while the corresponding DR-anchored recurrent estimators exhibit substantially larger tracking errors and higher divergence rates. The comparison indicates that the quality of the analytic estimator underlying the recurrent correction has a strong influence on closed-loop robustness. In particular, learning a residual correction on top of a well-conditioned EKF estimate is considerably more reliable than applying the same general recurrent-estimation strategy to dead reckoning.

\begin{figure}[t]
\centering
\includegraphics[width=0.92\linewidth]{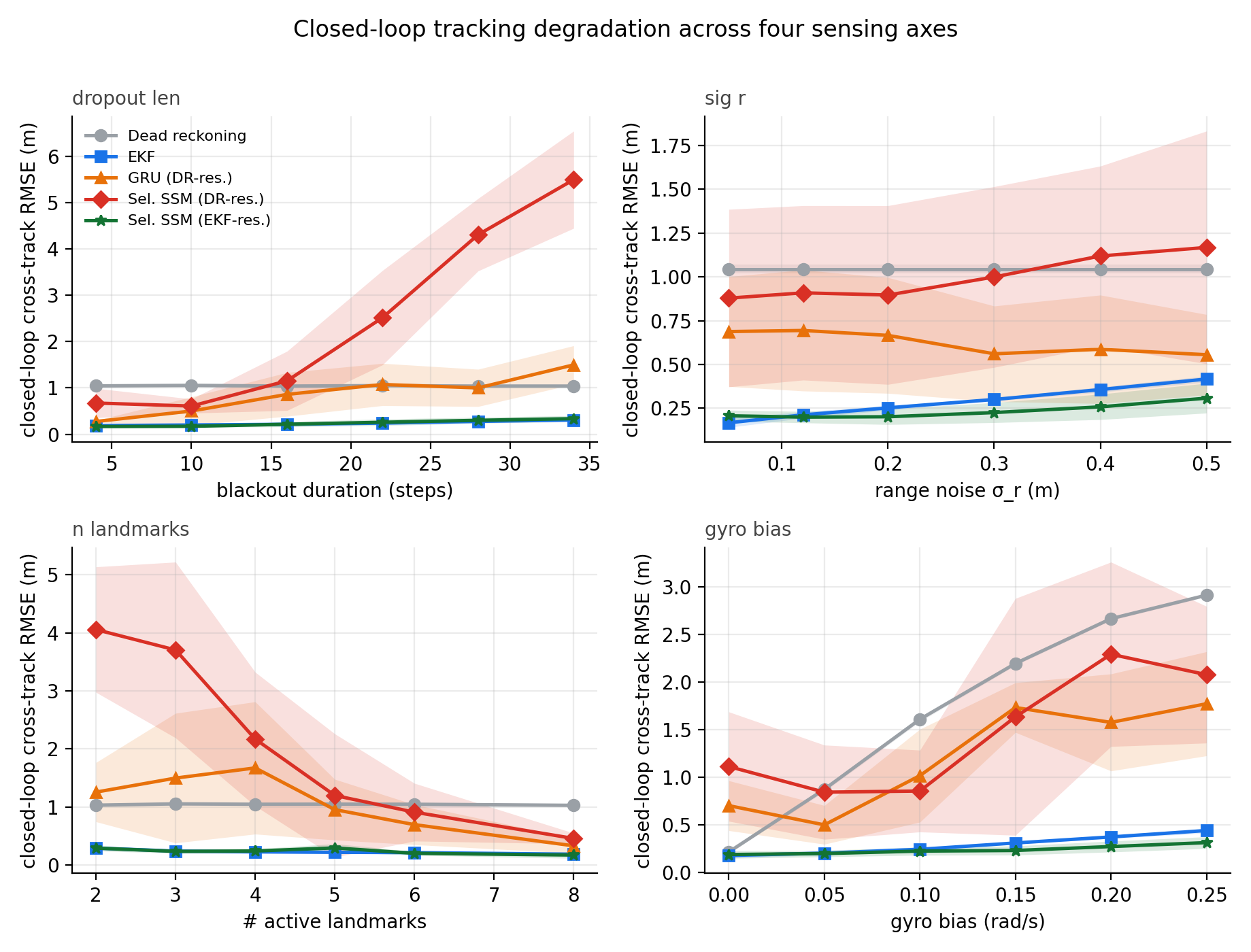}
\caption{Closed-loop cross-track RMSE across four sensing-degradation sweeps. The observer with the lowest tracking error changes with the sensing condition, while the EKF-anchored estimators remain the most consistently stable group.}
\label{fig:sweeps}
\end{figure}

Figure~\ref{fig:sweeps} extends this comparison across the four sensing-degradation sweeps. The observer with the lowest tracking error changes as sensing quality changes, although the EKF and the two EKF-anchored recurrent estimators remain the most stable group over most operating conditions. During the longest complete sensing outages, the standard EKF can outperform the learned EKF residual observers. In those intervals, no new landmark information is available, so the recurrent correction cannot rely on fresh exteroceptive measurements and must propagate from its previous internal state together with the odometry stream.

Across all 144 observer-condition pairs, replay position RMSE has a Spearman rank correlation of $\rho=0.923$ with closed-loop cross-track RMSE. The overall association is therefore strong, but the rankings are not identical. In 5 of the 24 sensing conditions, the observer selected by replay position RMSE is different from the observer with the lowest closed-loop tracking error. This result establishes the first distinction between offline estimation accuracy and feedback performance: even when the two quantities are strongly correlated in aggregate, they need not identify the same estimator for a particular operating condition.

\subsection{Measurement-free rollout error is less predictive of closed-loop performance}
\label{sec:imagination}

A multi-step rollout might be expected to provide a better indication of closed-loop performance than conventional replay because it directly evaluates how the estimator propagates its state over time. Our results show that this expectation depends strongly on the measurement-update pattern used during the rollout.

\begin{table}[t]
\centering
\small
\caption{Comparison of offline estimator-selection criteria across 24 sensing conditions and six observers. Higher rank correlation indicates better agreement with the closed-loop ordering; lower selection failures and regret indicate better estimator selection.}
\label{tab:selection}
\begin{tabular}{lcccc}
\toprule
Offline score & Spearman $\rho$ & Top-1 failures & Mean regret (m) & Max regret (m) \\
\midrule
Position RMSE (replay)        & \textbf{0.923} & \textbf{5/24}  & \textbf{0.0051} & \textbf{0.0347} \\
Heading RMSE (replay)         & 0.881 & 19/24 & 0.0237 & 0.0474 \\
Measurement-free RMSE, $H{=}20$  & 0.774 & 18/24 & 0.0265 & 0.1208 \\
\bottomrule
\end{tabular}
\end{table}

Table~\ref{tab:selection} compares the three offline selection criteria. The 20-step measurement-free rollout has a lower rank correlation with closed-loop cross-track RMSE than replay position error, with $\rho=0.774$ compared with $\rho=0.923$. It also selects an observer other than the closed-loop optimum in 18 of the 24 sensing conditions, compared with 5 of 24 for replay position RMSE. The corresponding maximum selection regret increases from $0.0347$~m for replay position RMSE to $0.1208$~m for the measurement-free rollout.

\begin{figure}[t]
\centering
\includegraphics[width=0.98\linewidth]{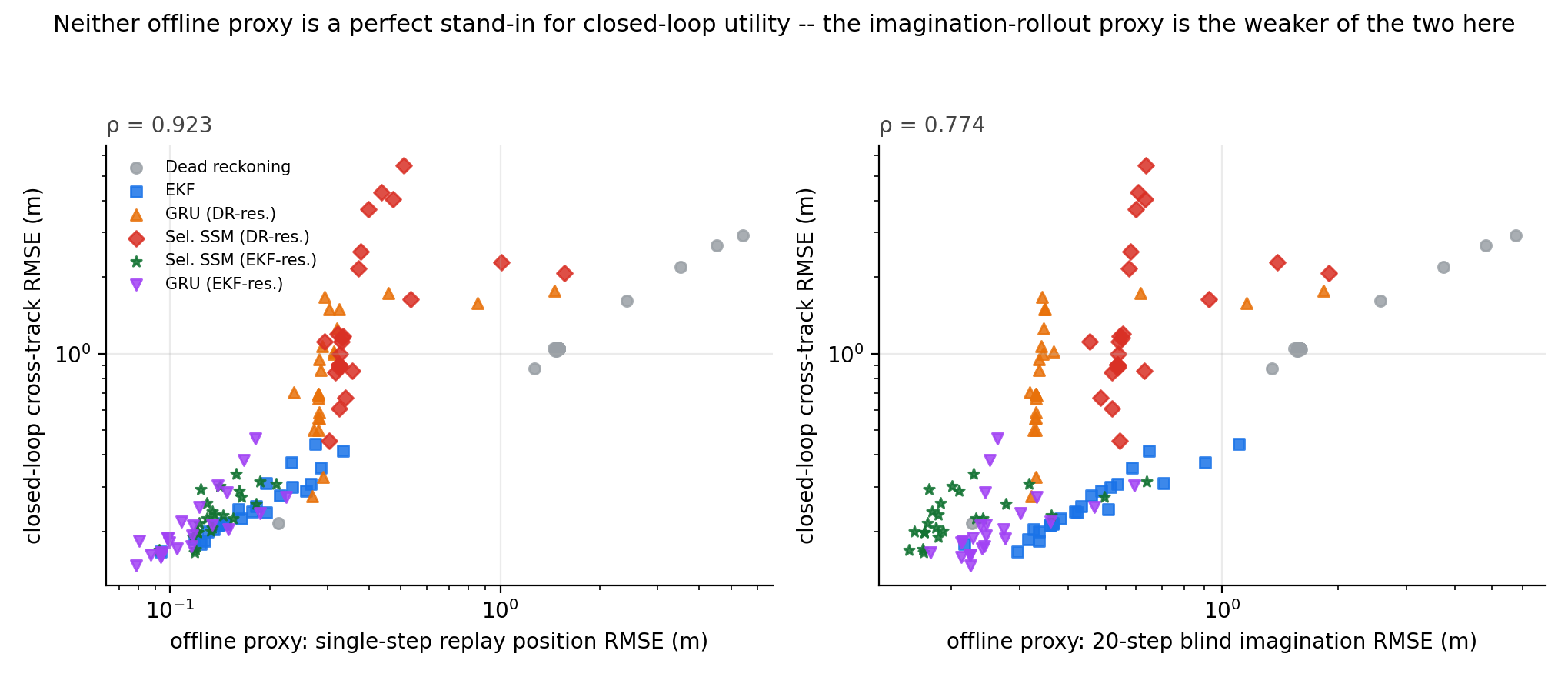}
\caption{Closed-loop cross-track RMSE versus two offline estimator scores for the same 144 observer-condition pairs. Replay position RMSE (left) is more closely associated with closed-loop tracking performance than the 20-step measurement-free rollout error (right).}
\label{fig:proxy-scatter}
\end{figure}

Figure~\ref{fig:proxy-scatter} shows the same relationship over the complete set of observer-condition pairs. Replay position error follows the closed-loop tracking trend more closely, whereas the measurement-free rollout produces a broader spread of errors for observers that later behave similarly in feedback. The longer rollout therefore contains useful information about open-loop state propagation, but that information is not sufficient to preserve the closed-loop estimator ranking.

The two evaluations measure different properties of the estimator. During the measurement-free rollout, the observer is asked to propagate its state for two seconds without receiving landmark corrections. Closed-loop operation follows a different sequence: state propagation, intermittent measurement updates, and feedback control are repeatedly interleaved. An estimator that accumulates noticeable drift during an extended measurement outage may still recover effectively when the next landmark measurement arrives. Conversely, an estimator can exhibit relatively small open-loop propagation error while producing corrections that lead to less favorable closed-loop tracking once those estimates are supplied to the controller.

\begin{figure}[t]
\centering
\includegraphics[width=0.90\linewidth]{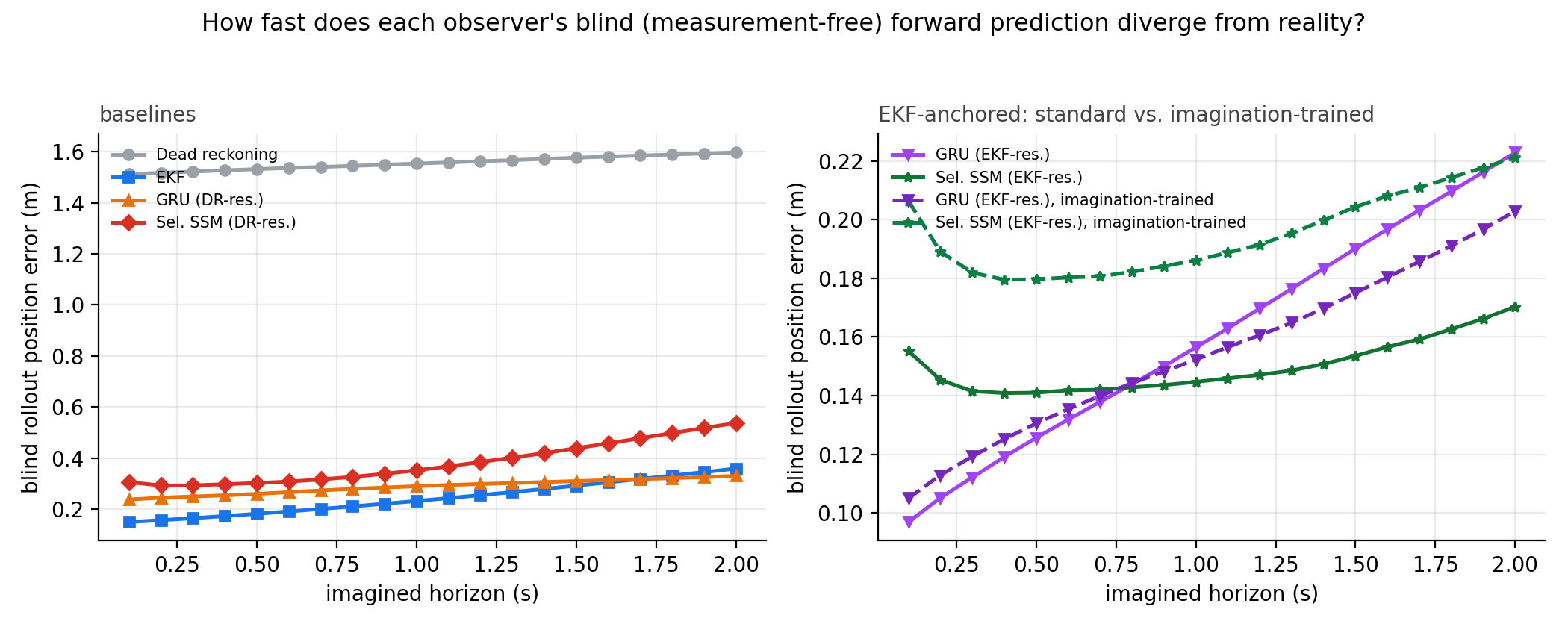}
\caption{Position error as the measurement-free rollout horizon increases in the nominal condition. Left: the six primary observers. Right: standard and long-blackout-trained EKF-anchored observers. Lower measurement-free propagation error does not necessarily correspond to lower closed-loop tracking error.}
\label{fig:imagination-growth}
\end{figure}

This distinction is also visible in Figure~\ref{fig:imagination-growth}. The rate at which position error grows during measurement-free propagation differs considerably across observers, yet that ordering does not reproduce the ordering obtained from closed-loop path tracking. These results suggest that rollout duration alone is not sufficient to characterize the relevance of an offline prediction test. The measurement-update schedule during the rollout must also be considered.

\subsection{Training with longer sensing outages provides condition-dependent improvements}
\label{sec:curriculum}

We next examine whether exposing recurrent estimators to longer intervals without landmark measurements during training improves their subsequent closed-loop performance. The standard and long-blackout-trained ($\dagger$) versions of SSM-EKF, GRU-EKF, and SSM-DR are evaluated under both combined sensing degradation and the isolated blackout-duration sweep.

\begin{table}[t]
\centering
\small
\caption{Closed-loop performance under combined sensing degradation consisting of high wheel slip, high gyroscope bias, high range noise, sparse landmark availability, and a long sensing outage. Values are averaged over ten evaluation seeds. $\dagger$ denotes training with extended sensing outages.}
\label{tab:compound}
\begin{tabular}{lcc}
\toprule
Observer & Cross-track RMSE (m) & Relative change \\
\midrule
EKF               & 1.800 & - \\
SSM-EKF           & 1.936 & - \\
SSM-EKF$^\dagger$ & 1.419 & $-27\%$ \\
GRU-EKF           & 1.717 & - \\
GRU-EKF$^\dagger$ & \textbf{1.061} & $\mathbf{-38\%}$ \\
SSM-DR            & 5.290 & - \\
SSM-DR$^\dagger$  & 5.411 & $+2\%$ \\
\bottomrule
\end{tabular}
\end{table}

Under the combined sensing-degradation condition, extended-blackout training improves both EKF-anchored recurrent estimators. SSM-EKF decreases from $1.936$~m to $1.419$~m cross-track RMSE, while GRU-EKF decreases from $1.717$~m to $1.061$~m (Table~\ref{tab:compound}). The same training modification does not improve the DR-anchored SSM. Its cross-track RMSE changes from $5.290$~m to $5.411$~m, indicating that longer periods of measurement-free training are not sufficient to compensate for the substantially weaker dead-reckoning reference estimate.

\begin{figure}[t]
\centering
\includegraphics[width=0.94\linewidth]{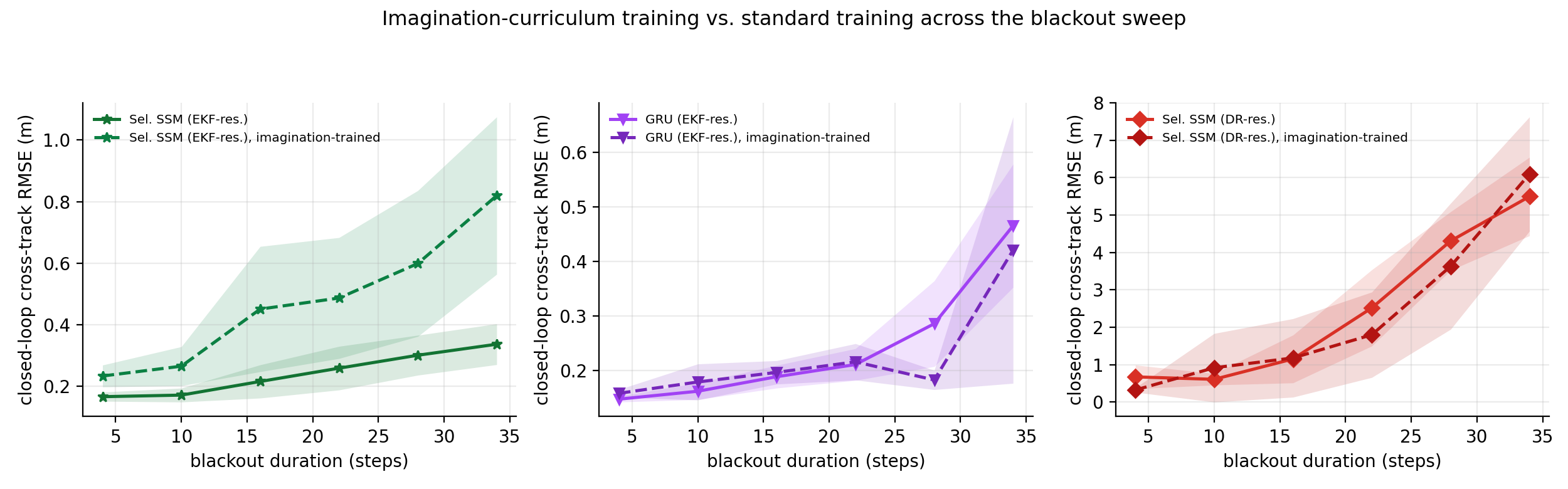}
\caption{Closed-loop cross-track RMSE for standard and extended-blackout training across the isolated sensing-outage sweep. GRU-EKF benefits primarily at the longest outages, whereas SSM-EKF degrades across the full sweep and SSM-DR shows no consistent improvement.}
\label{fig:curriculum}
\end{figure}

The isolated sensing-outage sweep provides a more detailed view of this behavior (Figure~\ref{fig:curriculum}). GRU-EKF$^\dagger$ remains close to the standard GRU-EKF through most of the sweep and becomes clearly better only for the two longest outages. At a blackout duration of 28 steps, for example, its cross-track RMSE is $0.183$~m compared with $0.286$~m for the standard GRU-EKF.

SSM-EKF responds differently to the same training modification. SSM-EKF$^\dagger$ has higher tracking error than the standard SSM-EKF at every tested blackout duration. The difference increases from $0.233$ versus $0.166$~m at a four-step outage to $0.819$ versus $0.336$~m at 34 steps. Its measurement-free rollout error is also higher throughout this sweep despite the additional exposure to long sensing outages during training. Three independently trained SSM-EKF$^\dagger$ models preserve the same ordering at the longest blackout condition, as reported in Appendix~\ref{app:seeds}.

The effect of longer-blackout training is therefore dependent on both estimator structure and sensing condition. The improvement observed for the EKF-anchored models under combined degradation does not translate into a uniform improvement under isolated sensing outages, and the same training procedure affects the GRU and SSM residual models differently. These results indicate that robustness to prolonged sensing loss cannot be inferred from the blackout duration represented in the training data alone. The quality of the underlying analytic estimate and the manner in which the recurrent model uses that estimate remain important.

\subsection{Separating rollout horizon from measurement-update frequency}
\label{sec:reconditioning}

The previous experiment leaves an important question unresolved. The weaker agreement obtained from the $H=20$ measurement-free rollout could arise because the prediction horizon is too long, because measurement updates are withheld, or because both effects interact. We therefore vary these two quantities independently.

We define the measurement-update interval $k$ as the number of prediction steps between landmark-based corrections during the offline rollout. A value of $k=1$ incorporates available measurements at every step, while $k=\infty$ corresponds to the measurement-free rollout used in Section~\ref{sec:imagination}. We first hold the rollout horizon fixed at $H=20$ and vary
$k \in \{1,2,5,10,20,\infty\}$.

\begin{figure}[t]
\centering
\includegraphics[width=0.94\linewidth]{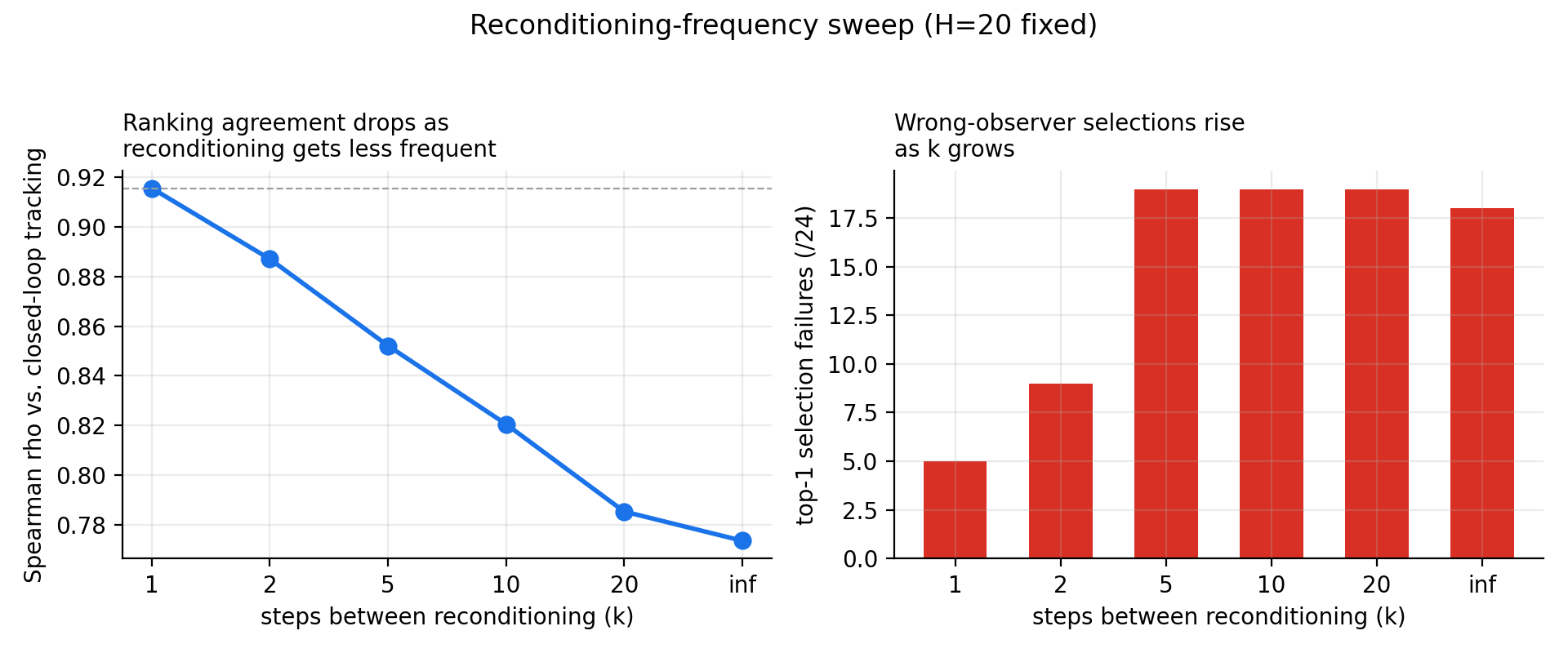}
\caption{Effect of the measurement-update interval for a fixed rollout horizon of $H=20$. The interval $k$ is varied over $\{1,2,5,10,20,\infty\}$ across the same 24 sensing conditions and six primary observers. Agreement with the closed-loop ranking decreases as measurement corrections become less frequent.}
\label{fig:reconditioning}
\end{figure}

Figure~\ref{fig:reconditioning} shows a systematic decrease in agreement with closed-loop performance as the interval between measurement corrections increases. Spearman rank correlation decreases from $\rho=0.916$ at $k=1$ to $\rho=0.774$ when measurements are withheld for the complete rollout. The number of top-1 selection failures increases from 5 of 24 conditions at $k=1$ to a peak of 19 of 24 conditions at $k=5$, $10$, and $20$, settling at 18 of 24 once measurements are withheld entirely, which is exactly the measurement-free rollout result in Table~\ref{tab:selection}. With the rollout horizon held fixed, this experiment isolates the effect of the measurement-update schedule and shows that regular state correction improves the ability of the offline rollout to reproduce the closed-loop observer ranking.

\begin{figure}[t]
\centering
\includegraphics[width=0.62\linewidth]{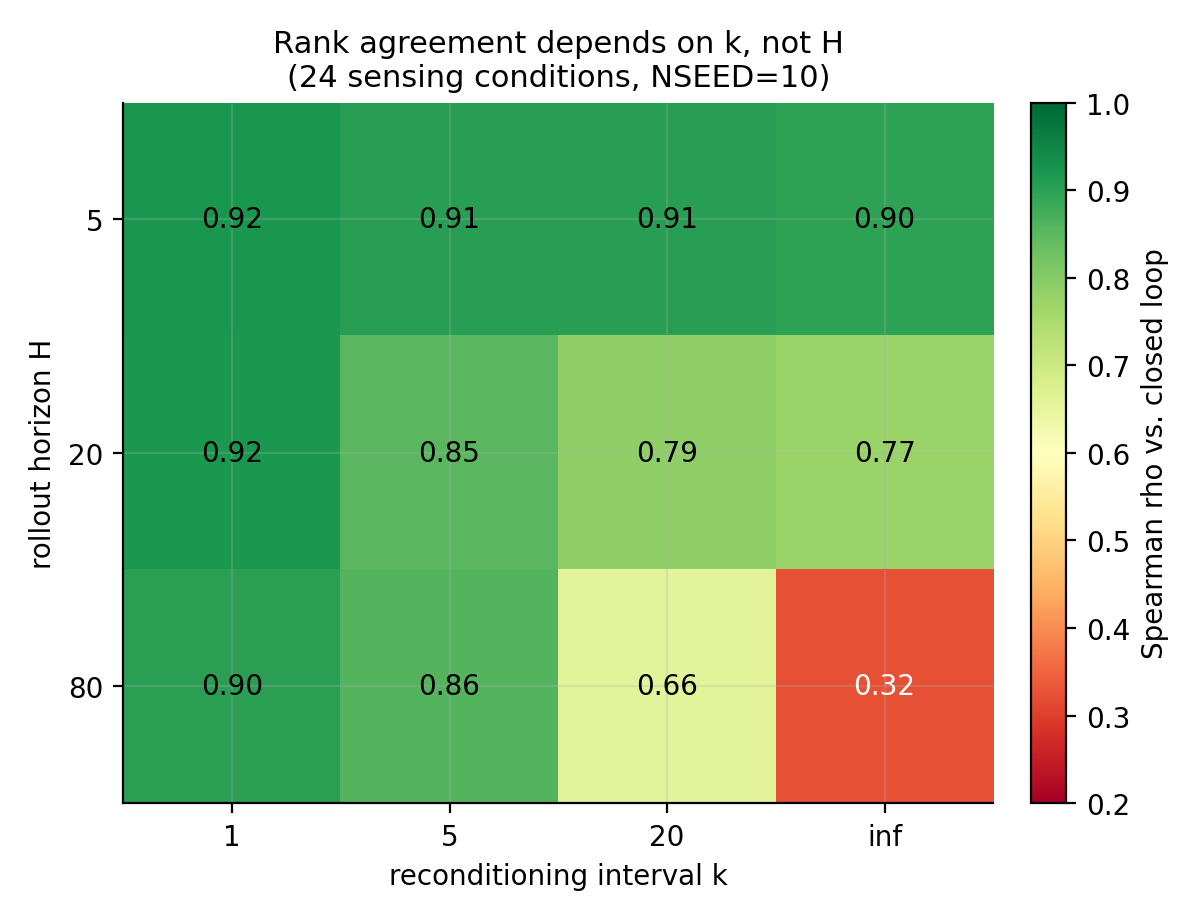}
\caption{Spearman rank correlation with closed-loop cross-track RMSE over a grid of rollout horizons $H \in \{5,20,80\}$ and measurement-update intervals $k \in \{1,5,20,\infty\}$, evaluated across all 24 sensing conditions and ten evaluation seeds. Long horizons retain relatively high ranking agreement when frequent measurement corrections are available, while the combination of long propagation and no correction produces the largest disagreement.}
\label{fig:horizongrid}
\end{figure}

We next vary both the rollout horizon and the measurement-update interval. Figure~\ref{fig:horizongrid} reports Spearman correlation over
$H \in \{5,20,80\}$ and
$k \in \{1,5,20,\infty\}$,
evaluated across all 24 sensing conditions and ten evaluation seeds, the same protocol used throughout the paper.

The two-dimensional sweep shows that rollout horizon and measurement-update frequency interact rather than acting independently. When measurements are incorporated at every step ($k=1$), rank correlation remains high as the horizon increases, with $\rho=0.92$, $0.92$, and $0.90$ for $H=5$, 20, and 80, respectively. Similarly, at the shortest horizon ($H=5$), correlation remains between $0.92$ and $0.90$ across the full range of update intervals because there is relatively little time for propagation error to accumulate.

The largest degradation occurs when a long prediction horizon is combined with infrequent or absent measurement correction. At $H=80$ and $k=\infty$, rank correlation falls to $\rho=0.32$, all 24 evaluated conditions select an observer different from the closed-loop optimum, and the mean selection regret reaches $0.76$~m. In contrast, the same $H=80$ horizon retains $\rho=0.90$ when measurements are incorporated at every step.

These results clarify the earlier measurement-free rollout result. Increasing the prediction horizon does not, by itself, cause an offline evaluation to lose correspondence with closed-loop behavior. The larger mismatch appears when prediction is extended over many steps without the measurement updates that would normally be available to the estimator during operation. For predictive state models used within a feedback loop, rollout horizon and measurement-update frequency should therefore be specified and evaluated jointly.

\section{Implications for World-Model Evaluation in Robotics}

Although the experimental platform considered in this study is intentionally simple, the underlying evaluation issue applies more broadly to predictive models used in robotic systems. The information available during an offline rollout often differs from the information available when the same model is used online. An offline benchmark may evaluate a long prediction from a fixed initial context, whereas a deployed robot may receive new camera images, proprioceptive measurements, force measurements, or other sensor updates after only a few control cycles. The state estimate is then corrected and the controller is evaluated again from the updated state. As a result, the usefulness of a predictive model depends not only on how accurately it propagates the system forward, but also on the sensing and feedback schedule under which those predictions are used.

\textbf{Evaluation should reproduce the measurement-update schedule of the intended application.}
A rollout horizon $H$ does not fully describe an evaluation protocol unless the availability of measurements during that horizon is also specified. For a robotic system that receives frequent sensor updates and replans at every control cycle, an evaluation based entirely on measurement-free propagation tests a different operating condition from the one encountered during normal use. In contrast, a robot expected to operate for extended periods without external measurements should be evaluated with correspondingly long prediction intervals. The appropriate rollout protocol therefore depends on both the prediction horizon and the frequency with which the model receives new measurements and updates its state.

\textbf{Prediction accuracy should be reported together with closed-loop agreement.}
State-prediction error remains an important measure of model quality, but it does not directly describe how estimation errors affect the controlled system. Offline evaluation can therefore be strengthened by reporting quantities that compare model rankings at the prediction and control levels, such as estimator-selection accuracy, policy-ranking agreement, closed-loop regret, or task success. In our experiments, the 20-step measurement-free rollout provides a meaningful measure of state propagation, yet it produces a substantially different estimator ranking from the one obtained in closed-loop path tracking. Reporting rank agreement and selection regret makes this distinction visible even when average state-prediction errors appear reasonable.

\textbf{State propagation and measurement correction should be evaluated separately.}
Predictive models used in robotics increasingly combine multiple sensing modalities, including vision, proprioception, force sensing, and other external measurements. The arrival of a new measurement changes the state estimate and can substantially alter the subsequent control action. A model that accumulates drift during a long period of open-loop propagation may still perform well if it incorporates the next measurement effectively. Conversely, a model with accurate short-term propagation can perform poorly if its state corrections introduce errors that are unfavorable for feedback control. Evaluation should therefore characterize both how the model propagates the system between measurements and how its state estimate evolves when new measurements become available.

These observations do not imply that short-horizon evaluation is generally preferable to long-horizon prediction. Predictive models used for planning, model-predictive control, or policy learning may rely on substantially longer periods of model propagation than the state estimators considered here. In those applications, long-horizon prediction accuracy may be directly relevant to performance. Our results instead indicate that the relationship between rollout accuracy and closed-loop utility depends on how the predictive model is incorporated into the robotic system. Evaluation protocols should therefore reproduce the sensing, state-update, and control structure of the intended application as closely as possible.

\section{Limitations}

The present study was designed to isolate the interaction between state prediction, measurement correction, and feedback control, and this controlled design necessarily limits the range of robotic systems represented in the experiments. The robot state is three-dimensional and does not include image observations or a learned visual representation. The primary experiments use one simulated differential-drive platform, one reference-path family, and a pure-pursuit controller. Appendix~\ref{app:policy} extends the comparison to a second, time-indexed controller and includes a policy-in-the-loop rollout in which the predicted state directly determines subsequent control inputs. The study does not yet include a learned action-generation model, a visual predictive model, or a second dynamical platform such as a quadrotor operating with intermittent position measurements.

The amount of training replication also differs across experiments. The four principal EKF-anchored recurrent estimators: SSM-EKF, GRU-EKF, and their extended-blackout counterparts are each evaluated using three independent training seeds, as reported in Appendix~\ref{app:seeds}. Five or more independent training runs would provide a stronger estimate of training variability, particularly for the SSM-EKF model trained with extended sensing outages, which exhibits noticeably larger variation across initializations. Some secondary experiments involving DR-anchored observers and ablations use fewer independent training runs and should therefore be interpreted primarily as supporting evidence rather than as independent statistical conclusions.

The matched-data study in Appendix~\ref{app:matched} controls for training-set size while changing the distribution of sensing-outage durations. In that experiment, matching the training blackout distribution to the evaluation conditions does not provide a consistent closed-loop advantage. The result should therefore be interpreted as evidence that blackout-duration matching alone is insufficient, rather than as evidence for a particular training curriculum.

Several additional experiments remain deliberately limited in scope. The combined sensing-degradation experiment represents one difficult operating condition rather than a full factorial combination of all disturbance levels. The controlled error-shape study in Appendix~\ref{app:errorshape} uses a shadow-controller rollout to isolate the effect of different estimation-error structures while holding their position RMSE fixed; this provides a clear mechanism-level comparison but does not replace evaluation in the full closed-loop simulator. The broader project also includes localization experiments using physical MRCLAM sensor logs, but those data were collected from an existing robot trajectory and therefore cannot provide the closed-loop validation required to test the central feedback claim of this paper.

These limitations define the scope of the conclusions. The experiments do not establish a universal evaluation metric for large visual world models, nor do they claim that measurement-free rollouts are inappropriate for robotic planning in general. They demonstrate a more specific result: when a predictive state model is used inside a feedback controller and receives intermittent measurement corrections, an offline rollout can produce a misleading model ranking if its measurement schedule differs substantially from the one encountered during closed-loop operation. Extending the same controlled comparison to visual predictive models, additional robot dynamics, different controllers, and physical closed-loop experiments is an important direction for future work.

\section{Conclusion}

This paper examined how offline prediction accuracy relates to closed-loop control performance when a predictive state estimator is used inside a robotic feedback loop. In a differential-drive path-tracking task with biased odometry and intermittent landmark sensing, we compared replay-based estimation error, multi-step measurement-free rollout error, and closed-loop tracking performance across six observers and 24 sensing conditions. The 20-step measurement-free rollout was less consistent with the closed-loop observer ranking than replay position RMSE, despite explicitly evaluating multi-step state propagation.

The horizon and measurement-update experiments explain this result more precisely. Long prediction horizons remain informative when the estimator continues to receive regular measurement corrections, while the correspondence with closed-loop performance deteriorates substantially when long propagation is combined with infrequent or absent measurement updates. The relevant evaluation variable is therefore not prediction horizon alone, but the combination of prediction horizon and measurement-update schedule.

The training experiments lead to a similar conclusion. Exposure to longer sensing outages improves the EKF-anchored recurrent estimators under combined sensing degradation, including a reduction in GRU-EKF cross-track RMSE from $1.72$~m to $1.06$~m. The improvement is not uniform across isolated sensing outages or estimator architectures, and matching the training outage distribution to the evaluation conditions does not by itself produce better closed-loop behavior. Robustness therefore depends on the interaction between the analytic estimator, the learned correction, and the operating regime rather than on blackout duration alone.

Taken together, the results suggest a practical principle for evaluating predictive models in robotics. Prediction error should be measured under an information pattern that reflects how the model will actually be used. For models operating inside a feedback controller, this means specifying when new measurements become available, when the state estimate is updated, and when new control inputs are computed. A useful rollout benchmark should therefore reproduce the sensing and feedback structure of the intended robotic application, rather than treating prediction horizon as the sole measure of temporal difficulty.

\section*{Acknowledgments}
This project began as a final project for ECE~6562 (Autonomous Control of Robotic Systems), Georgia Institute of Technology, Summer 2026, and was extended with the blind-rollout evaluation and long-blackout training experiments reported here.

\bibliographystyle{plainnat}
\bibliography{references}

\appendix

\section{Sensitivity to Training Initialization}
\label{app:seeds}

\begin{figure}[h]
\centering
\includegraphics[width=0.92\linewidth]{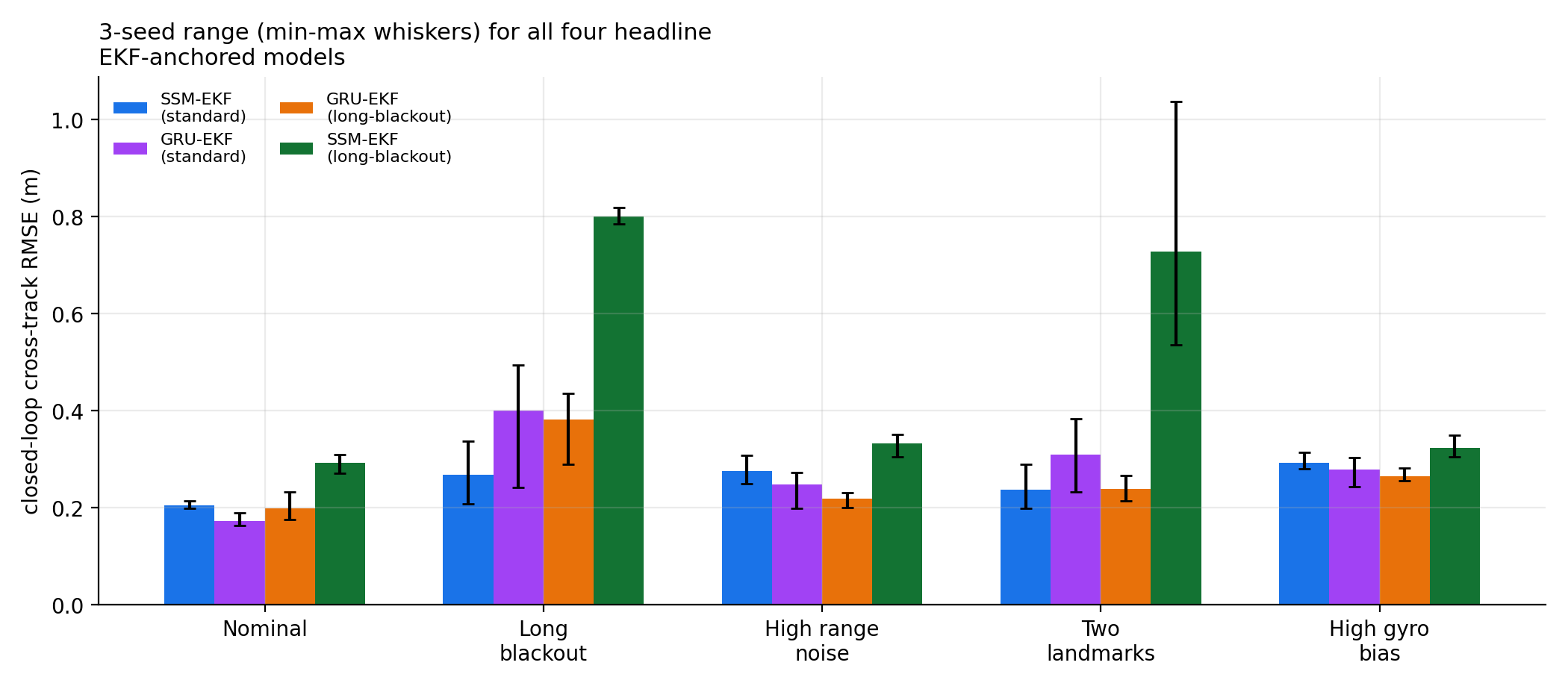}
\caption{Closed-loop cross-track RMSE across three training seeds for the four principal EKF-anchored recurrent estimators. Whiskers indicate the minimum and maximum values across seeds. SSM-EKF$^\dagger$ exhibits the largest sensitivity to initialization, particularly under the longest sensing outage and with only two visible landmarks.}
\label{fig:seedvariance}
\end{figure}

The main results involve recurrent estimators whose performance can vary with training initialization. We therefore evaluate the four principal EKF-anchored models: SSM-EKF, GRU-EKF, and their extended-outage variants using three independent training seeds. Table~\ref{tab:seeds-ssmekf} reports the individual results for SSM-EKF$^\dagger$, while Figure~\ref{fig:seedvariance} summarizes the corresponding variation for all four EKF-anchored models.

\begin{table}[h]
\centering
\small
\caption{Cross-track RMSE (m) for SSM-EKF$^\dagger$ across three independent training seeds. The degradation under the longest sensing outage is reproduced across all three initializations.}
\label{tab:seeds-ssmekf}
\begin{tabular}{lccc}
\toprule
Condition & seed 0 & seed 1 & seed 2 \\
\midrule
Nominal             & 0.309 & 0.295 & 0.271 \\
Long blackout (34)  & 0.819 & 0.799 & 0.785 \\
High range noise    & 0.305 & 0.351 & 0.340 \\
Two landmarks       & 1.037 & 0.536 & 0.611 \\
High gyro bias      & 0.316 & 0.349 & 0.305 \\
\bottomrule
\end{tabular}
\end{table}

The repeated runs support two observations. First, the increased tracking error of SSM-EKF$^\dagger$ under the longest sensing outage is not associated with a single unfavorable initialization. All three independently trained models produce cross-track RMSE above $0.78$~m, whereas the standard SSM-EKF remains substantially lower under the same condition. Second, the amount of training variability depends on the training procedure itself. The extended-outage SSM-EKF exhibits considerably larger variation across seeds than the other EKF-anchored models, particularly when landmark information is sparse. This result reinforces the need to evaluate recurrent estimators across multiple training initializations when comparing robustness under degraded sensing.

\section{Policy-Coupled Rollout Evaluation}
\label{app:policy}

The measurement-free rollout used in Section~\ref{sec:imagination} propagates each observer using the realized future odometry sequence from the recorded trajectory. This construction isolates state-prediction error, but the predicted state does not influence the subsequent control inputs during the offline rollout. We therefore introduce a second rollout protocol in which the predicted state is coupled directly to the controller.

In this policy-coupled rollout, the current predicted pose is supplied to the path-tracking controller, the controller computes the next input from that pose, and the simulated state is propagated using the resulting control command. The updated predicted state is then used to compute the following command, and the procedure is repeated for the complete rollout horizon. Future control inputs are therefore determined by the model's own predicted trajectory rather than replayed from the recorded run.

To make the offline and closed-loop computations directly comparable, this experiment uses a time-indexed pure-pursuit controller for both the trajectory generation and the offline rollout. We compare conventional replay, a periodically corrected rollout with $k=5$, the measurement-free rollout, and the policy-coupled rollout over the same 24 sensing conditions and six primary observers.

\begin{table}[h]
\centering
\small
\caption{Agreement between four offline evaluation protocols and the corresponding closed-loop observer ranking. Results use 24 sensing conditions, six primary observers, and ten evaluation seeds. The numerical values differ from Table~\ref{tab:selection} because this experiment uses a time-indexed controller for both trajectory generation and offline rollout.}
\label{tab:rollout-modes}
\begin{tabular}{lcccc}
\toprule
Mode & Spearman $\rho$ & Top-1 failures & Mean regret (m) & Max regret (m) \\
\midrule
Replay                              & \textbf{0.853} & 18/24 & 0.0354 & 0.1249 \\
Measurement-updated ($k=5$, $H{=}20$) & 0.799 & \textbf{17/24} & 0.0302 & 0.0950 \\
Measurement-free ($H{=}20$)        & 0.796 & \textbf{15/24} & \textbf{0.0224} & \textbf{0.0768} \\
Policy-coupled ($H{=}20$)          & 0.743 & 20/24 & 0.0479 & 0.1439 \\
\bottomrule
\end{tabular}
\end{table}

\begin{figure}[h]
\centering
\includegraphics[width=0.92\linewidth]{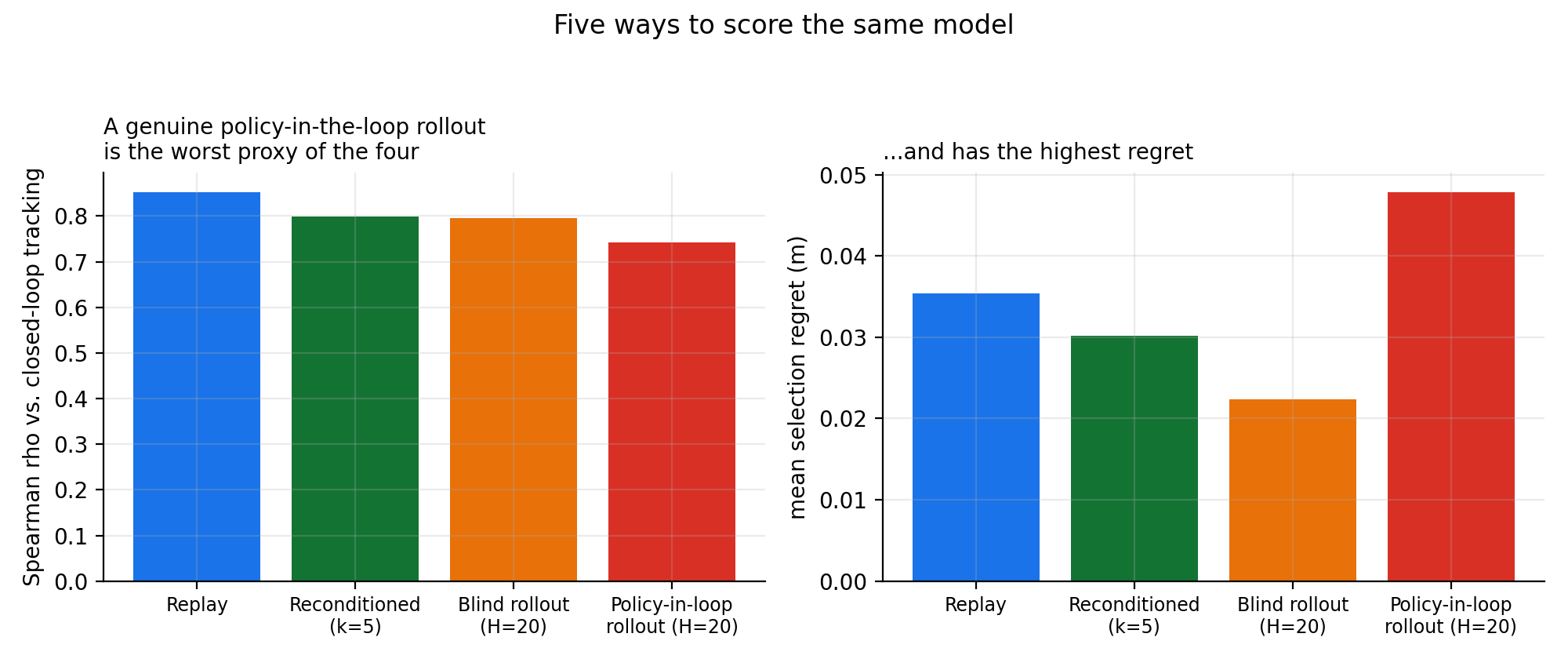}
\caption{Comparison of offline evaluation protocols. The policy-coupled rollout has the lowest rank correlation with closed-loop performance and the largest mean selection regret among the four protocols considered.}
\label{fig:rolloutmodes}
\end{figure}

Table~\ref{tab:rollout-modes} shows that coupling the predicted state back into the controller does not improve agreement with the closed-loop observer ranking. The policy-coupled rollout has the lowest Spearman correlation, $\rho=0.743$, and the largest mean selection regret, $0.0479$~m. Its performance is also worse than the measurement-free rollout in rank correlation and mean regret.

The result can be understood from the propagation of estimation error through feedback. In the measurement-free rollout, estimator error affects the predicted state while the future odometry sequence remains fixed. In the policy-coupled rollout, the same state error also changes the next control input, which changes the subsequent predicted trajectory and therefore the control inputs that follow. The offline rollout consequently accumulates both state-prediction error and controller-induced trajectory deviation without receiving the measurement corrections available during normal operation.

This experiment strengthens the distinction between reproducing a long autonomous rollout and reproducing the information structure of the deployed feedback system. A rollout can include the controller explicitly and still provide a poor estimate of closed-loop model ranking when the sensing and measurement-update schedule differs from that of the actual robot.

\section{Effect of Estimation-Error Structure on Control}
\label{app:errorshape}

Position RMSE measures the magnitude of translational error but does not distinguish between different temporal patterns or directions of that error. A feedback controller can respond very differently to errors with the same overall RMSE. We therefore construct six synthetic estimation-error sequences to isolate this effect. Five sequences are normalized to the same position RMSE of $0.30$~m, while the sixth contains only a heading bias and has zero position error by construction.

Each error sequence is injected into the same reference trajectory using a counterfactual controller rollout. At every step, the controller receives the ground-truth pose perturbed by the prescribed estimation error. The resulting control command is then applied to nominal vehicle dynamics. This construction keeps the reference trajectory, controller, and overall position-error magnitude fixed while changing only the temporal and directional structure of the estimation error.

\begin{figure}[h]
\centering
\includegraphics[width=0.85\linewidth]{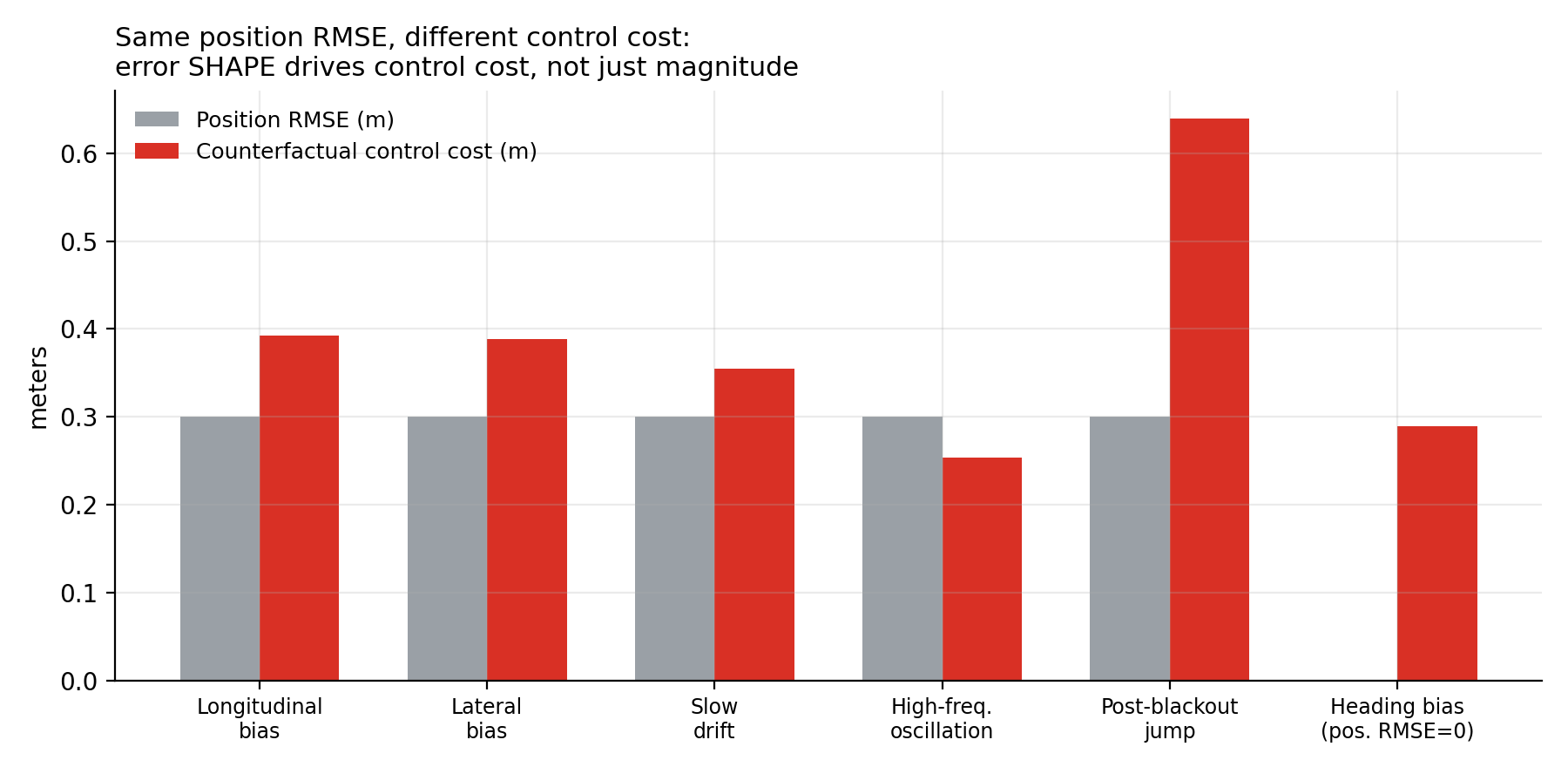}
\caption{Closed-loop effect of synthetic estimation-error sequences. Five sequences have identical position RMSE of $0.30$~m, yet their resulting tracking cost ranges from $0.25$~m for high-frequency oscillatory error to $0.64$~m for a persistent post-outage displacement. A pure heading bias has zero position RMSE but produces $0.29$~m of tracking error.}
\label{fig:errorshapes}
\end{figure}

Figure~\ref{fig:errorshapes} shows that equal position RMSE does not imply equal control performance. The persistent post-outage displacement produces $0.64$~m of tracking error, approximately $2.5\times$ the $0.25$~m produced by the high-frequency oscillatory sequence despite the two having identical position RMSE. The controller partially attenuates the rapidly alternating error, whereas a persistent displacement continues to bias the commanded trajectory over many control cycles.

The heading-only experiment provides a second illustration of the same issue. Although its position RMSE is zero by construction, the heading bias produces $0.29$~m of tracking error because orientation error directly changes the steering command. These examples provide a mechanism for the ranking differences observed in the main experiments. Measurement outages and subsequent measurement corrections can produce errors with different temporal structures for example, gradual drift during an outage followed by a sharp correction when sensing resumes and a scalar pose RMSE does not distinguish how those structures interact with the controller.

\section{Controlling for Training-Set Size and Outage Distribution}
\label{app:matched}

The extended-outage models in Section~\ref{sec:curriculum} are trained on 300 trajectories, compared with 200 trajectories for the original training set. The initial comparison therefore changes both the amount of training data and the distribution of sensing-outage durations. To separate these effects, we train two additional 300-trajectory variants while keeping the architecture, optimizer, batch size, and initialization procedure unchanged.

The first variant, \textbf{Standard-300}, uses the original sensing distribution with 300 trajectories and therefore changes only the amount of training data. The second, \textbf{Deployment-matched-300}, samples outage durations directly from the evaluation levels
$\{4,10,16,22,28,34\}$. These models are compared with \textbf{Long-blackout-300}, which uses the extended outage distribution introduced in Section~\ref{sec:curriculum}.

\begin{figure}[h]
\centering
\includegraphics[width=0.92\linewidth]{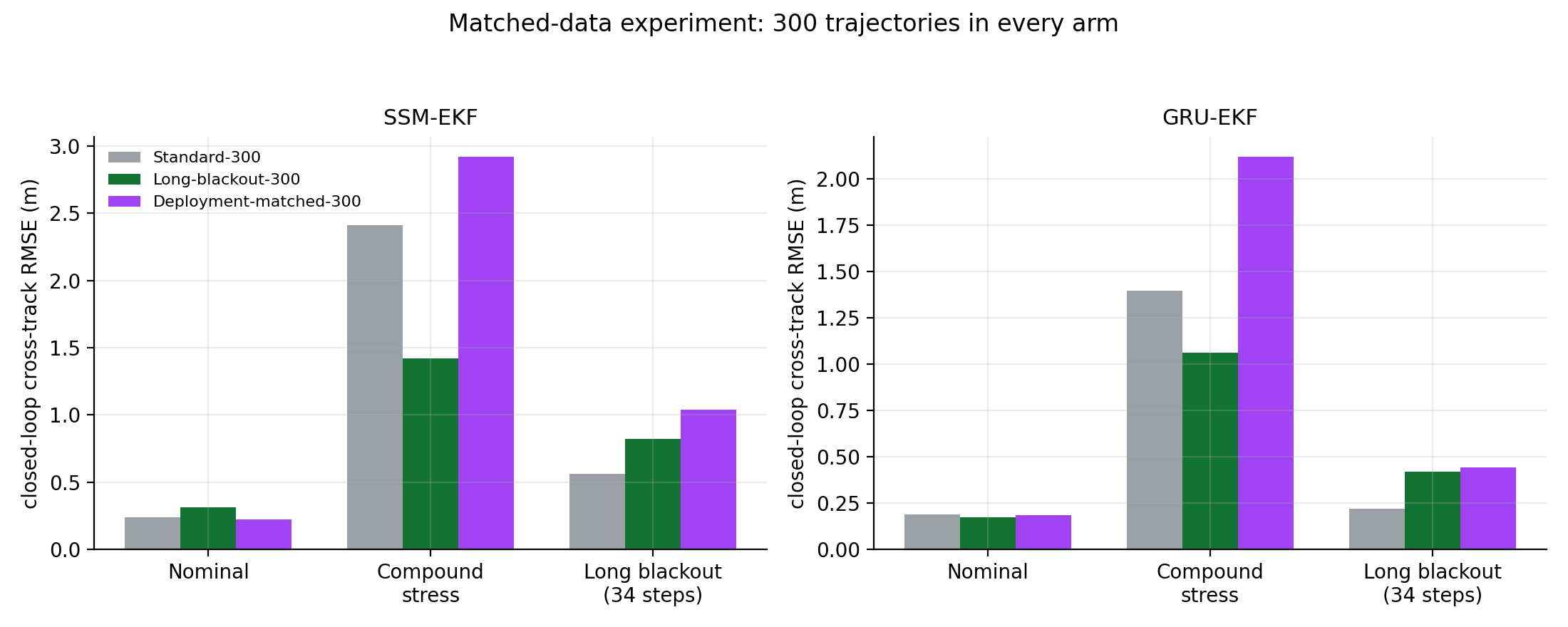}
\caption{Closed-loop cross-track RMSE for three 300-trajectory training distributions under nominal sensing, combined sensing degradation, and an isolated long sensing outage. Extended-outage training performs best under combined degradation for both EKF-anchored recurrent architectures, while directly matching the evaluation outage distribution does not provide a consistent advantage.}
\label{fig:matcheddata}
\end{figure}

Figure~\ref{fig:matcheddata} shows that the improvement observed under combined sensing degradation cannot be attributed solely to the larger training set. For SSM-EKF, Long-blackout-300 achieves $1.42$~m cross-track RMSE under combined degradation, compared with $2.41$~m for Standard-300 and $2.92$~m for Deployment-matched-300. The same ordering appears for GRU-EKF, with corresponding errors of $1.06$, $1.40$, and $2.12$~m.

The isolated long-outage condition produces a different ordering. Standard-300 gives the lowest tracking error for both recurrent architectures, with $0.56$~m for SSM-EKF and $0.22$~m for GRU-EKF. Directly matching the training outage durations to the evaluation sweep therefore does not provide a consistent closed-loop advantage.

Taken together, these results suggest that the benefits of extended-outage training are associated with the interaction between training distribution, estimator architecture, and sensing condition rather than with training-set size or outage-duration matching alone. A training distribution that improves robustness under several simultaneous sensing degradations can still reduce performance under an isolated sensing outage.

\section{Condition-Level Ranking Analysis and Statistical Significance}
\label{app:stats}

The pooled Spearman correlations reported in Table~\ref{tab:selection} are computed over all 144 observer-condition pairs. Because dead reckoning performs substantially worse than the EKF-based estimators under many conditions, this pooled statistic may partly reflect the separation between clearly different observer families. We therefore perform an additional condition-level analysis in which the ranking statistics are computed separately for each sensing condition and then summarized across conditions.

Table~\ref{tab:percondition} reports the median and minimum per-condition Spearman correlation, pairwise ranking accuracy, top-1 selection accuracy, and CVaR-20\% selection regret. Pairwise accuracy measures the fraction of observer pairs that an offline metric orders in the same way as the closed-loop evaluation. CVaR-20\% regret reports the mean selection regret over the five highest-regret conditions and therefore emphasizes the tail of the selection-error distribution.

\begin{table}[h]
\centering
\small
\caption{Condition-level ranking statistics across 24 sensing conditions and six primary observers. CVaR-20\% regret is the mean selection regret over the five highest-regret conditions.}
\label{tab:percondition}
\begin{tabular}{lccccc}
\toprule
Offline score & Median $\rho$ & Min $\rho$ & Pairwise acc. & Top-1 acc. & CVaR20 regret (m) \\
\midrule
Replay position RMSE        & \textbf{0.943} & \textbf{0.657} & \textbf{0.903} & \textbf{0.792} & \textbf{0.0244} \\
Replay heading RMSE         & 0.771 & 0.600 & 0.803 & 0.208 & 0.0429 \\
Measurement-free ($H{=}20$)& 0.800 & 0.429 & 0.808 & 0.250 & 0.0590 \\
\bottomrule
\end{tabular}
\end{table}

Replay position RMSE provides the strongest agreement with closed-loop behavior across all of the condition-level statistics in Table~\ref{tab:percondition}. Its median per-condition Spearman correlation is $0.943$, and it correctly orders $90.3\%$ of observer pairs. Its CVaR-20\% regret is $0.0244$~m, compared with $0.0590$~m for the 20-step measurement-free rollout. The difference therefore remains visible even when the analysis is performed independently within each sensing condition rather than over the pooled observer-condition set.

We additionally test whether the difference in estimator-selection regret between replay position RMSE and the measurement-free rollout can be explained by variation across the 24 sensing conditions. A paired bootstrap is performed by resampling sensing conditions with replacement while preserving the pairing between the two evaluation metrics. We also perform a paired sign-flip permutation test on the condition-level regret differences. The observed difference in mean regret,
\[
    R_{\mathrm{replay}} - R_{\mathrm{rollout}} = -0.0214~\mathrm{m},
\]
has a 95\% bootstrap confidence interval of
$[-0.033,-0.011]$~m and a sign-flip permutation value of
$p=0.0001$.

The same analysis is applied to the two endpoints of the measurement-update sweep in Section~\ref{sec:reconditioning}. Comparing $k=1$ with $k=\infty$ gives a mean regret difference of $-0.0214$~m, with a 95\% confidence interval of $[-0.033,-0.011]$~m and $p=0.0001$. This is the same regret difference as the replay-versus-rollout comparison above because, condition by condition, $k=1$ and $k=\infty$ select the same observers as replay position RMSE and the measurement-free rollout, respectively. These results show that the observed differences in selection regret persist when the comparison is performed condition by condition rather than being driven only by the pooled ranking statistic.

\section{Code and reproducibility}
All simulation, training, and evaluation code, bundled checkpoints, and raw result JSON files are included in the GitHub repository.
\end{document}